\documentclass[10pt,twocolumn,letterpaper]{article}

\usepackage[pagenumbers]{cvpr} 

\usepackage{times}
\usepackage{epsfig}
\usepackage{graphicx}
\usepackage{amsmath}
\usepackage{amssymb}
\usepackage{dsserif}

\usepackage[table]{xcolor}

\usepackage{makecell}
\usepackage{bbding}

\usepackage{graphicx}
\usepackage{subcaption}
\usepackage{float}
\usepackage{caption}
\usepackage{lscape}                                         
\usepackage{wrapfig}

\usepackage[lined,ruled,linesnumbered]{algorithm2e}
\usepackage{animate} 

\usepackage{booktabs}                   
\usepackage{multirow}
\usepackage{arydshln} %

\usepackage{enumitem}

\usepackage{bm}                          

\usepackage{amssymb}
\usepackage{amsmath}  

\usepackage{units}
\usepackage{color}
\usepackage[T1]{fontenc}    
\usepackage{amsfonts}       
\usepackage[utf8]{inputenc} 
\usepackage{nicefrac}       
\usepackage{microtype}      
\usepackage{pifont}         
\usepackage{comment}

\usepackage[mathscr]{euscript}
\usepackage{colortbl}

\usepackage{url}  

\newlength\paramargin

\newcommand{\mfigure}[2]
{
\includegraphics[width=#1\linewidth]{#2}
}

\newcommand{\heading}[1]
{
\vspace{1mm}\noindent\textbf{#1}
}

\newcommand{\secref}[1]{Section~\ref{sec:#1}}
\newcommand{\figref}[1]{Figure~\ref{fig:#1}} 
\newcommand{\tabref}[1]{Table~\ref{tab:#1}}

\long\def\ignorethis#1{}

\newbox\jsavebox%

\graphicspath{{figure}, {example}}

\usepackage{xcolor}
\colorlet{dark-blue}{blue!50!black}
\colorlet{dark-cyan}{cyan!75!black}
\colorlet{dark-purple}{purple!50!black}
\colorlet{dark-red}{red!75!black}
\colorlet{dark-green}{green!75!black}
\colorlet{dark-orange}{orange!50!black}
\colorlet{dark-gray}{black!75}
\colorlet{light-gray}{black!30}
\definecolor{nice-red}{HTML}{E41A1C}
\definecolor{nice-orange}{HTML}{FF7F00}
\definecolor{nice-yellow}{HTML}{FFC020}
\definecolor{nice-green}{HTML}{39b54a}
\definecolor{nice-blue}{HTML}{0071bc}
\definecolor{nice-purple}{HTML}{984EA3}

\colorlet{verylight-gray}{black!10}
\definecolor{LightCyan}{rgb}{0.88,1,1}

\definecolor{best}{rgb}{1, 0.85, 0.7}
\definecolor{second}{rgb}{1,1, 0.8}

\definecolor{cvprblue}{rgb}{0.21,0.49,0.74}
\usepackage[pagebackref,breaklinks,colorlinks,citecolor=cvprblue]{hyperref}

\makeatletter
\def\@fnsymbol#1{\ensuremath{\ifcase#1\or \dagger\or \ddagger\or
		\mathsection\or \mathparagraph\or \|\or **\or \dagger\dagger
		\or \ddagger\ddagger \else\@ctrerr\fi}}
\makeatother

\makeatletter
\def\thanks#1{\protected@xdef\@thanks{\@thanks
		\protect\footnotetext{#1}}}
\makeatother

\graphicspath{{figures/}}

\hypersetup{
	citecolor=dark-green,
	filecolor=blue,
	linkcolor=nice-red,
	urlcolor=nice-red
}

\usepackage[capitalize]{cleveref}
\crefname{section}{Sec.}{Secs.}
\Crefname{section}{Section}{Sections}
\Crefname{table}{Table}{Tables}
\crefname{table}{Tab.}{Tabs.}

\title{Motion Blur Decomposition with Cross-shutter Guidance}

\author{Xiang Ji \hspace{0.15in} 
	Haiyang Jiang \hspace{0.15in} 
	Yinqiang Zheng\footnotemark[1]  \thanks{$\dagger$ Corresponding author}  \thanks{\textcolor{magenta}{\url{https://github.com/jixiang2016/dualBR}}}\\
	The University of Tokyo, Japan\\
	\tt\small \{jixiang,jiang-haiyang777\}@g.ecc.u-tokyo.ac.jp,\\
	\tt\small yqzheng@ai.u-tokyo.ac.jp 
	\vspace{-2mm}\\
}

\begin{document}
\maketitle

\begin{abstract}

Motion blur is a frequently observed image artifact, especially under insufficient illumination where exposure time has to be prolonged so as to collect more photons for a bright enough image. Rather than simply removing such blurring effects, recent researches have aimed at decomposing a blurry image into multiple sharp images with spatial and temporal coherence. Since motion blur decomposition itself is highly ambiguous, priors from neighbouring frames or human annotation are usually needed for motion disambiguation. In this paper, inspired by the complementary exposure characteristics of a global shutter (GS) camera and a rolling shutter (RS) camera, we propose to utilize the ordered scanline-wise delay in a rolling shutter image to robustify motion decomposition of a single blurry image. To evaluate this novel dual imaging setting, we construct a triaxial system to collect realistic data, as well as a deep network architecture that explicitly addresses temporal and contextual information through reciprocal branches for cross-shutter motion blur decomposition. Experiment results have verified the effectiveness of our proposed algorithm, as well as the validity of our dual imaging setting. 

\end{abstract}

\vspace{-4mm}
\section{Introduction}
\label{sec:intro}

Photo capture usually needs sufficient exposure time to collect photons. If ego-motion of camera or dynamic objects are presented during this period, the resultant image will suffer from motion blur. This kind of degradation makes visual content less interpretable. Hence, extensive research has been devoted to reverse this process and produce sharper details. 

\begin{figure}[!tb]
	\centering
	{
			\mfigure{1}{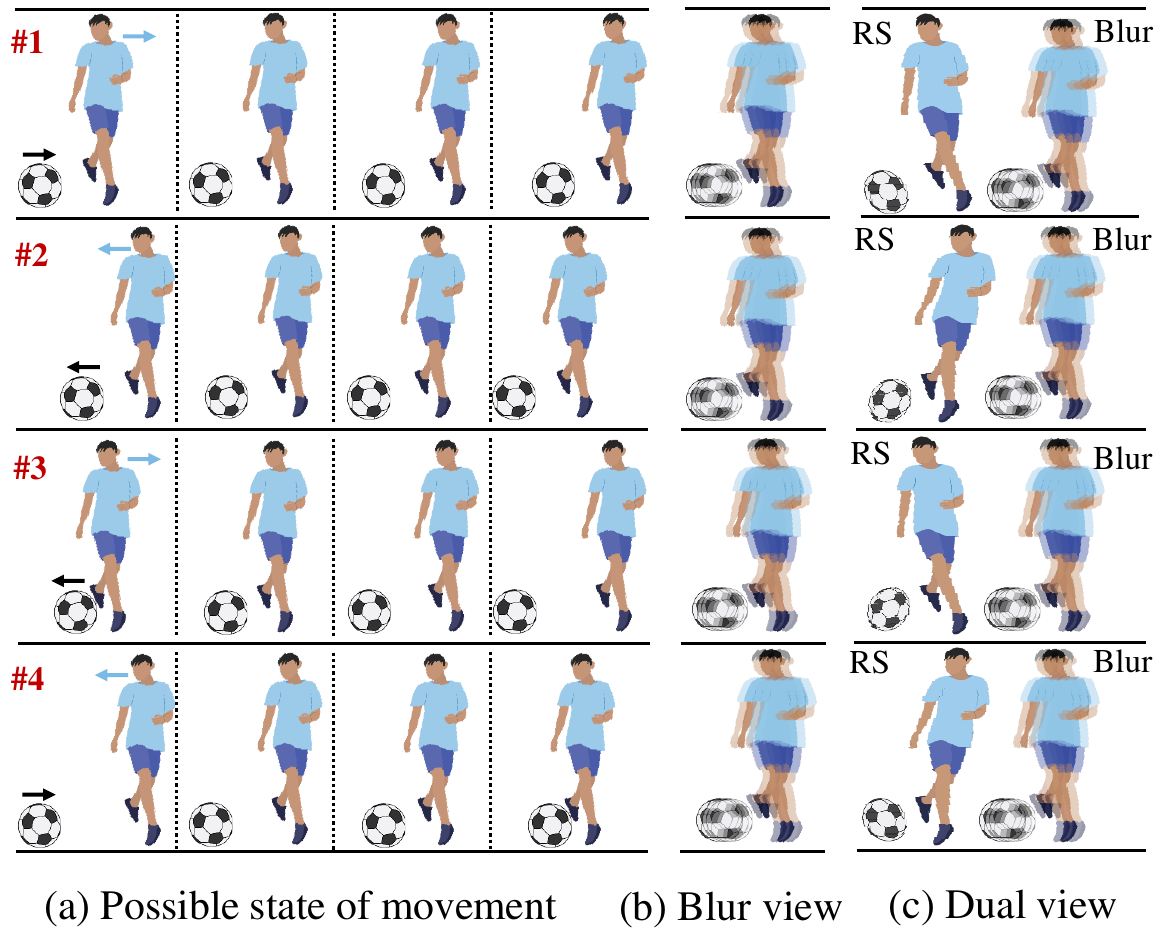}
	}
	\vspace{-6mm}
	\caption{
		\textbf{Motion ambiguity of blur observation.} In this toy example, we show two objects: a soccer and a player, both moving horizontally. (a) shows four possible motion states (both moving right, both moving left, one moving left and the other is towards right.) during the exposure time. (b) presents corresponding motion blurred observations. 
		They are all identical due to averaging effects, which brings about motion ambiguity to blur decomposition. (c) In our dual Blur-RS setting, rolling shutter (RS) view implicitly encoded temporal ordering of latent frames.
	}
	\label{fig:motion_ambiguity}
	\vspace{-3mm}
\end{figure}

Generally, the deblurring course is formulated as image-to-image transition by extracting a single latent frame from a blurry input~\cite{nah2017deep,park2020multi,purohit2019region,tao2018scale,wang2022uformer,gao2019dynamic}. Recently, researchers step froward to a more ambitious task, retrieving a sharp image sequence instead of just one frame, dubbed as blur decomposition~\cite{purohit2019bringing}. Unfortunately, averaging effects of motion blur have severely destroyed the temporal ordering of latent frames, which cannot be restored solely by reconstruction loss~\cite{jin2018learning}. To make matters worse, motion ambiguity resides in each dynamic object, thus leading to numerous plausible solutions, yet many of which are physically infeasible. ~\figref{motion_ambiguity} (a) and (b) illustrate this ambiguity by using two dynamic objects. For a given blur observation, there exist four motion sequences that could be interpreted as its decomposition. Models only trained on this data usually have issues of instability and low performance~\cite{zhong2022animation}.

Handling the loss of temporal order is a problem far from being well-studied in the blur decomposition. Current solutions mainly fall into two categories: (a) introducing ordering-invariant loss~\cite{jin2018learning} and (b) approximating latent temporal order within exposure by motion of consecutive blur frames~\cite{zhong2022animation,zhong2022blur}. The former one is easily caught in sub-optimal solutions owing to the weak supervision propagated from the loss. Similarly, the latter solution suffers from severe degeneration when presented with long exposure time or fast motion. Moreover, motion estimation among blurry frames is inherently nontrivial and time-consuming. Recently, it has been recognized that rolling shutter (RS) images encode the canceled motion due to its row-by-row exposure mode~\cite{fan2021inverting,fan2022context}, according to which RS effects can be mitigated. But restoring a sequence from single RS input still remains unfinished because of lacking complete global content. In contrast, Blur images contain adequate contextual information but without temporal ordering. On the other hand, dual camera system has been widely exploited in RS correction (RS-RS, RS-Event)~\cite{albl2020two,zhong2022bringing,zhou2022evunroll}, deblurring (GS-Event) ~\cite{sun2022event,xu2021motion}, even vibration sensing (GS-RS)~\cite{sheinin2022dual}.

Therefore, considering the complementarity of Blur and RS images, we propose dual Blur-RS setting to solve the motion ambiguity of blur decomposition. As shown in ~\figref{motion_ambiguity} (c), the RS view not only provides local details but also implicitly captures temporal order of latent frames. Meanwhile, GS view could be exploited to mitigate the initial-state ambiguity from RS counterpart (as discussed in ~\secref{setting_justification}). Inspired by the hardware design of ~\cite{rim2020real,zhong2021towards}, we devised our triaxial imaging system to capture strictly aligned high-speed sharp videos and low-speed Blur-RS pair videos.

 Facilitated by the collected dataset, we further proposed a novel two-staged model, containing motion interpretation and blur decomposition modules, to reconstruct a sharp video sequence from cross shutter views: blur and RS observations. The motion interpretation module firstly disentangles the bilateral motion fields as complementary dual stream: Blur and RS branches, to actively address the contextual characterization and temporal abstraction, respectively. Shutter alignment and aggregation enables mutual boosting of two branches by propagating aligned and aggregated feature to each other. Besides, the temporal positional encoding further enhances model's ability to disambiguate motion direction of latent frames. Subsequently, estimated motion fields along with blur-RS inputs will be warped and refined through blur decomposition module to generate a sharp video clip. At last, we also deeply explored the advantages of our proposed Blur-RS combination over existing settings for blur and RS decomposition by providing experimental results. In summary, our main contributions are:
\begin{itemize}
	\item We present a new setting of dual Blur-RS combination to address the motion ambiguity of blur decomposition, and demonstrate its superiority to pure RS or blur setting.

    \item Rather than a biaxial system for image-to-image deblurring, we develop a triaxial imaging system that simultaneously captures Blur-RS pairs along with high-speed ground truth, and collect a real dataset named RealBR.

    \item We introduce a novel neural network architecture that actively address the contextual characterization and temporal abstraction by dual stream motion interpretation module. Extensive experiments have validated the effectiveness of our setting and model.
\end{itemize}

\section{Related Work}
\label{sec:relate}

\subsection{Blur Decomposition}
 Compared with traditional deblurring task, reconstructing an image sequence from single blurred input is much more challenging because the average effects have destroyed the temporal ordering of latent frames. Jin \emph{et al}\onedot ~\cite{jin2018learning} raise this problem and address the temporal ordering ambiguity for the first time. From two aspects, they design a network with large receptive field to tackle the inherent ill-posedness of deblurring and ordering-invariant loss for motion ambiguity. ~\cite{purohit2019bringing} specially proposes a surrogate task to learn motion representation from sharp videos in an unsupervised manner, and then employs it as a guidance of training motion encoder for blurred images. ~\cite{argaw2021restoration} takes advantages of spatial transformer network modules to restore a video sequence and its underlying motion in an end-to-end manner. In order to avoid the directional ambiguity, BiT ~\cite{zhong2022blur} takes three consecutive blurry frames as input to extract the motion prior. They propose a blur intra-interpolation transformer based on novel multi-scale Swin transformer blocks along with dual-end temporal supervision and symmetric ensembling strategies to effectively unravel the underlying motion within the exposure time and recover arbitrary sharp frames from blur. Zhong \emph{et al}\onedot ~\cite{zhong2022animation} also emphasize on the motion ambiguity of blur decomposition by introducing motion guidance representation. They provide three interfaces to acquire the motion guidance: learning by network, approximating from blur video and user input. The learning one is also supervised by optical flow of blur frames. So essentially, solutions proposed in ~\cite{purohit2019bringing,zhong2022blur,zhong2022animation} all try to represent the latent motion by using consecutive blurry inputs.
 
 Different from blur decomposition, another line of related work inherits video frame interpolation setting by substituting inputs as blurry frames to reconstruct clear images within deadtime. Most of them take as input an image sequence and interpolate a sharp one at the middle of two blurry frames~\cite{shen2020blurry,zhang2020video,jin2019learning}. The recent study has started to generate interpolated frame at arbitrary time~\cite{oh2022demfi,ji2023rethinking}. In this paper, we mainly focus on the motion ambiguity of blur decomposition. So, we do not expand the discussion of these works in detail.

\subsection{Dual Camera System}
Recently, significant progress has been made by constructing dual camera view to handle different vision tasks. According to combination manner, we briefly divide them as: RS-RS~\cite{albl2020two,zhong2022bringing}, RS-Event~\cite{zhou2022evunroll}, GS-Event~\cite{sun2022event,xu2021motion} and RS-GS~\cite{sheinin2022dual}.

Due to the ill-posedness of single RS correction, Albl \emph{et al}\onedot~\cite{albl2020two} resort to a camera configuration: two RS cameras with reversed scanning direction. They further proved that the setup possesses geometric constraints needed to correct rolling shutter distortion using only a sparse set of point correspondences between the two images. Lately, Zhong \emph{et al}\onedot~\cite{zhong2022bringing} extend this setup to learning based method. Instead of correcting RS geometrically, they develop an end-to-end model, to generate dual optical flow sequence through iterative learning. In contrast, considering the high-speed characteristic of the event camera, Zhou \emph{et al}\onedot~\cite{zhou2022evunroll} introduce a novel computational imaging setup consisting of an RS sensor and an event sensor to correct RS effects.

Similarly, by exploring the high-temporal resolution property of events, ~\cite{sun2022event,xu2021motion} use a cross-modal set up: GS-Event to solve the deblurring problem. Sun \emph{et al}\onedot~\cite{sun2022event} unfold blurring process into an end-to-end two stage restoration network by effectively fusing event and image features. They design an event-image cross-modal attention module, which allows network to focus on relevant features from the event branch and filter out noise. Xu \emph{et al}\onedot~\cite{xu2021motion} aim at data inconsistency, constructing a piece-wise linear motion model taking into account motion non-linearities, to achieve accurate deblurring in a self-supervised manner.

Most interestingly, Sheinin \emph{et al}\onedot~\cite{sheinin2022dual} simultaneously capture the vibration with two cameras equipped with rolling and global shutter sensors, respectively. The RS camera captures distorted speckle images that encode the high-speed object vibrations, while GS camera captures undistorted reference images of the speckle pattern, helping to decode the source vibrations.

\section{Methodology}
\label{sec:method}

\begin{figure*}[!tb]
	\centering
	\vspace{-1mm}
	\mfigure{1}{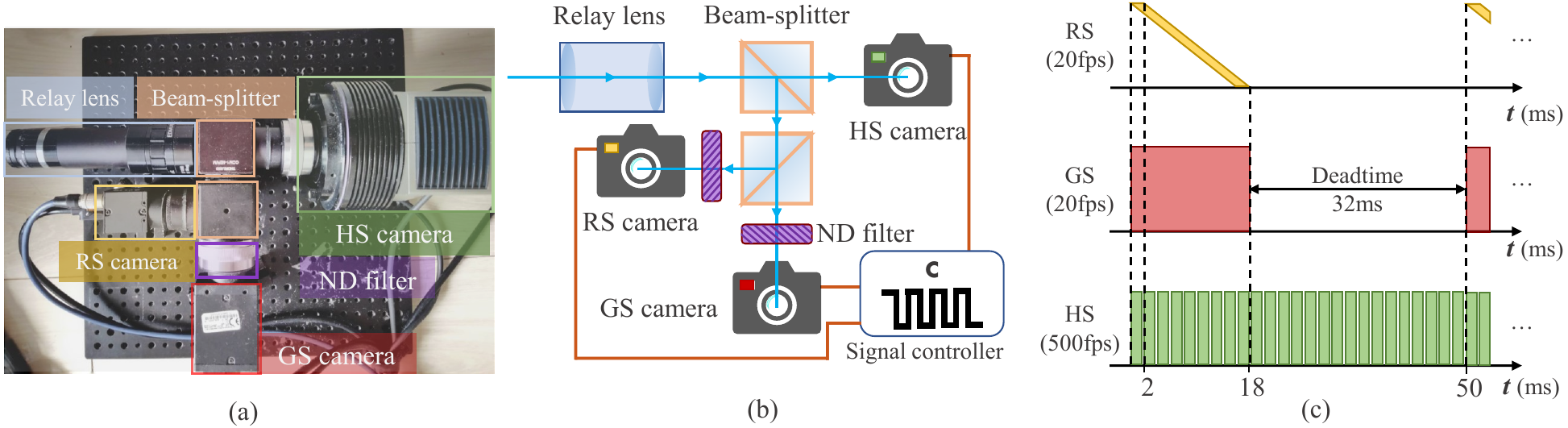}
	\vspace{-6mm}
	\caption{
		\textbf{Our triaxial imaging system.} (a) A photo of the actual system for data gathering; (b) Optical diagram of our system; (c) Illustration of exposure duration for all cameras on temporal axis. In picture (c), its vertical axes can be interpreted as spatial rows of captured images from each camera.
	}
	\label{fig:imaging_system}
	\vspace{-6mm}
\end{figure*}

\begin{table}[!tp]
	\centering
	\caption{\textbf{Specifications of our triaxial imaging system}. The deadtime between two adjacent high speed frames is extremely short and thus can be ignored.
	}
	\vspace{-3mm}
	\setlength\tabcolsep{4pt}	
	\resizebox{0.9\linewidth}{!}
	{
		\begin{tabular}{lccc}
			\toprule
			\textbf{Device} & \textbf{RS camera} & \textbf{GS camera} & \textbf{HS camera}\\
			\midrule
			{Resolution} & $800\!\times\!800$  & $800\!\times\!800$ & $800\!\times\!800$\\
			{Frame rate} & 20 fps & 20 fps & 500 fps\\  
			{Exp. per Row} & 2 ms  & 18 ms & 2 ms\\ 
			{Delay. per Row} & 20 $\mu$s  & 0 $\mu$s & 0 $\mu$s\\ 
			{Exp. per Frame} & 18 ms  & 18 ms & 2 ms\\
			{Deadtime} & 32 ms  & 32 ms  & 0 ms\\ 
			\bottomrule
		\end{tabular}
	}
	\label{tab:imaging_system}
	\vspace{-6mm}
\end{table}

\subsection{Proposed Setup of Blur Decomposition}
\label{sec:setup}

Usually, Blur accumulation can be formulated as an averaging process presented in a linear space by using inverse camera response function (CRF) on RGB images~\cite{nah2017deep,nah2019ntire}. While blur decomposition aims at extracting uniformly distributed sharp frames from single blurred image.


As we discussed in \secref{intro}, this process is highly ill-posed because of motion ambiguity residing in accumulation of photons. Considering the characteristic of RS exposure that inherently encodes temporal ordering of latent frames and provide local details as supplementary to global content of blur, we additionally capture an RS view $R$ of each blurred frame $B$ so as to better address the indeterminacy. As a result, the decomposition problem is formulated as:
\begin{equation}
	\begin{split}
		(B,R) \mapsto \textbf{I} = \{S^t, t \in 0,\cdots,N-1 \} 
	\end{split}
\end{equation}

\subsection{Optical System and Dataset}
\label{sec:data_collect}

\paragraph{Optical System} 
In order to capture aligned training inputs (RS, GS) and ground truth sequences that can be recorded by a high-speed camera (HS), we constructed a triaxial optical system as depicted in Figure~\ref{fig:imaging_system}. Similar to ~\cite{zou2021learning}, the system comprises 2 beam-splitters that partition incident light into 3 identical beams, and 3 cameras for RS (FLIR BFS-U3-63S4C with 2x2 binning), GS (FLIR GS3-U3-23S6C), HS (a high-speed GS camera, BITRAN CS-700C, with forced cooling) separately. Other specifications for 3 cameras are detailed in Table~\ref{tab:imaging_system}.The usage of a neutral density (ND) filter with roughly 20\% transmittance is for the purpose of counterbalancing excessive brightness in blurry GS images brought about by their relatively long exposure duration. While beam-splitters feeding light signals of the same scene to 3 cameras, and the other ND-filter equalizing illumination magnitude of blur and RS views, we further incorporated geometrical transforms for pixel-level alignment and synchronization signal control for simultaneity. For more details, please refer to our supplementary materials.

\vspace{-2mm}
\paragraph{RealBR Dataset} 
Applying the optical system to capture real world image sequences, we established a RealBR (GS Blur \& RS) dataset by recording 54 distinct street scenes containing ample amount of objects, like vehicles and pedestrians, and various camera motions. In each scene, we have 56 pairs of consecutive RS and GS blur frames, and 1400 corresponding sharp HS images.
As can be seen from Figure~\ref{fig:imaging_system}(c), capturing of single pair of Blur-RS frames has a period of 50ms, with GS and RS cameras finishing their exposure in 18ms leaving the rest 32ms in one period as deadtime, while HS camera exposing within 2ms at 500fps taking 25 frames in total within one period, 9 of which temporally located inside GS/RS exposure duration and the rest 16 inside deadtime. After necessary preprocessing, we reorganized entire dataset and split it into 40, 4, and 10 scenes for training, validation and test. All captured images are in both RGB and RAW format, and will be made publicly available, facilitating related potential exploration in the area.

\subsection{The Proposed Architecture}
\label{sec:model}
\begin{figure*}[!tb]
	\centering
	\mfigure{1}{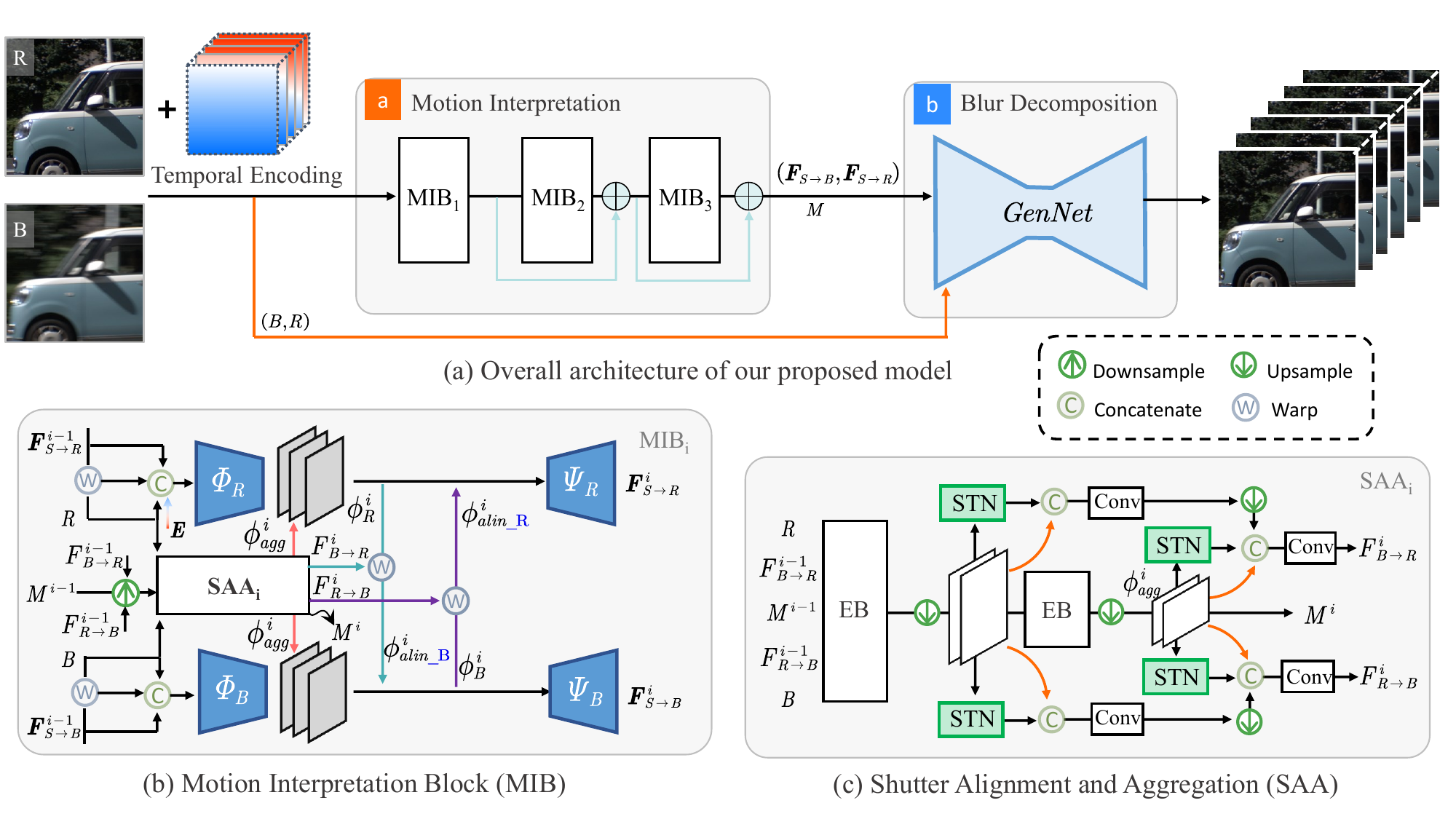}
	\vspace{-2mm}
	\caption{
		\textbf{Our proposed model.} (a) shows the overall architecture containing two stages: motion interpretation and blur decomposition. Blur decomposition is implemented through a \emph{GenNet}. Motion interpretation takes as input a blur image $B$ and an RS image $R$ along with its temporal positional encoding $E$. It consists of three blocks and one of them  is unfolded in (b). (c) presents specific details of shutter alignment and aggregation (SAA). Feature extracted by encoder block (EB) will be converted using spatial transformer network (STN), and then enhanced through a Conv. block to accurately predict displacement field between shutters.
	}
	\label{fig:model}
	\vspace{-4mm}
\end{figure*}

The overview of our proposed architecture is shown as \figref{model} (a). We mainly focus on reconstructing a clear video sequence from a blurred image with the assistance of its RS view to address the motion ambiguity issues. The inference process of our $\mathcal{F}$ is formulated as:
\begin{equation}
	\begin{split}
		\textbf{S} = \mathcal{F} \left(B, R \right)
	\end{split}
	\vspace{-1mm}
\end{equation}
where $\textbf{S} = \{S^t, t \in 0,\cdots,N-1 \} $ denotes extracted latent sharp video sequence with a length of $N$. $R$ is the RS view of blurred input $B$. 

Overall, the whole model could be divided into two stages: motion interpretation and blur decomposition. Motion interpretation (MI) is highly attentive to explore the benefits of our Blur-RS combination in an iterative manner with three motion interpretation blocks (MIB). With the guidance of temporal positional encoding, it explicitly emphasizes the contextual characterization and temporal abstraction from disentangled blur and RS streams, respectively. Estimating bidirectional motion fields can be described as bellow:
\begin{equation}
	\begin{split}
	( \textbf{F}_{S \rightarrow B}, \textbf{F}_{S \rightarrow R}, M ) = \mathcal{MI} \left(B, R \right)
	\end{split}
	\vspace{-1mm}
\end{equation}
where $ \textbf{F}_{S \rightarrow B} = \{ F_{S^t \rightarrow B}, t \in 1,\cdots,N \} $ are the intermediate flows from targeted latent frames to the blurry input. Similarly, $\textbf{F}_{S \rightarrow R}$ denotes counterparts from latent frames to the RS view and $M$ is a predicted mask to aggregate warped frames using bilateral motion fields. The blur decomposition part is implemented through \emph{GenNet}  in an encoder-decoder architecture to warp and refine the reconstructed latent video sequence, which is formulated as:
\begin{equation}
	\begin{split}
	\textbf{S} = GenNet \left(B, R, \textbf{F}_{S \rightarrow B}, \textbf{F}_{S \rightarrow R}, M \right)
	\end{split}
	\vspace{-1mm}
\end{equation}

Hereinafter, we highlight core components of the model: motion interpretation block with mutually boosted branches; temporal positional encoding; and shutter alignment and aggregation. As for the structure of \emph{GenNet} in blur decomposition, it is presented in supplemental materials.

\vspace{-2mm}
\paragraph{Dual Streams with Mutual Incentive}
As we observed, methods exploiting neighboring frames~\cite{zhong2022animation,zhong2022blur} or dual view from different cameras ~\cite{zhong2022bringing} tend to simply concatenate them to capture motion fields. But considering the fact regarding our cross shutter views that blur and RS inputs play contrasting roles in the blur decomposition task, we separately address them into two parallel branches. As shown in \figref{model} (b), the $i^{th}$ motion interpretation block (MIB$_i$) is implemented as two streams with mutual incentive through a shutter alignment and aggregation module. The RS branch offers local details and disambiguates motion directions. Meanwhile, blur counterpart with full global context will elevate the accuracy of motion magnitude and mitigate initial-state ambiguity residing in RS views.

To promote the interaction of two branches, the aligned and aggregated feature are extracted. Instead of directly concatenating encoded features from each other, we firstly predict bidirectional displacement maps between two input views: $F^{i}_{B \rightarrow R}$ and $F^{i}_{R \rightarrow B}$. Then corresponding aligned features $\phi^{i}_{alin\_R}$ and $\phi^{i}_{alin\_B}$ for two streams are warped as:
\begin{equation}
	\begin{split}
		\phi^{i}_{alin\_R} = \mathcal{W} \left(\phi^{i}_B, F^{i}_{R \rightarrow B} \right)\\
		\phi^{i}_{alin\_B} = \mathcal{W} \left(\phi^{i}_R, F^{i}_{B \rightarrow R} \right)
	\end{split}
	\vspace{-1mm}
\end{equation}
where $\mathcal{W}$ denotes backward-warping process. $\phi^{i}_B$ and $\phi^{i}_R$ are represented feature of blur and RS views. This aligning process enables us to adaptively selects helpful features and rejects incorrect ones from the other view. In addition, the aggregated feature $\phi^{i}_{agg}$ has also been taken into account as an auxiliary.

Therefore, taking the blur branch as an example, which can be formulated as:
\begin{equation}
	\begin{split}
		&\phi^{i}_B = \Phi_B \left(B, \textbf{F}^{i-1}_{S \rightarrow B} \right) \\
		&\textbf{F}^{i}_{S \rightarrow B} = \Psi_B \left( \left[ \phi^{i}_B, \phi^{i}_{alin\_B}, \phi^{i}_{agg} \right] \right)
	\end{split}
	\vspace{-1mm}
\end{equation}
where $\Phi_B $, $\Psi_B$ are encoder and decoder. $\left[ \cdot\right]$ denotes concatenation. The processing is similar under RS branch except for temporal positional encoding. Overall, the $i^{th}$ MIB can be described as:
\begin{equation}
    \begin{split}
        \textbf{F}^{i},  F^{i}, M^i = \mathcal{MIB}_i \left(B,R,\textbf{E},\textbf{F}^{i-1},F^{i-1}, M^{i-1} \right)
    \end{split}
    \vspace{-1mm}
\end{equation}
where $\textbf{F} = [\textbf{F}_{S \rightarrow B},\textbf{F}_{S \rightarrow R}]$, $ F =  [F_{B \rightarrow R}, F_{R \rightarrow B}] $ and $M$ is the predicted mask to aggregate warped frames in blur decomposition.

\vspace{-2mm}
\paragraph{Temporal Positional Encoding}
To further enhance model's ability to disambiguate motion direction of latent frames, we propose a temporal positional encoding for the RS branch in MIB. Row-by-row exposure of RS cameras inherently carves the latent motion into captured images. ~\cite{liu2020deep,fan2021sunet} approximates this process by copying image rows from corresponding latent frames to synthesize RS effects. So, naturally, the temporal positional encoding for RS input $R$ and latent frame $S^t$ are:
\begin{equation}
	\begin{split}
     [E_R]_k &= k, k=0,1,\cdots,N-1 \\
     E_{S^t} &= \frac{H-1}{N-1}t \cdot \mathbbb{1}
	\end{split}
	\vspace{-1mm}
\end{equation}
where $[\cdot]_k$ is the operation that extracts $k^{th}$ row and $\mathbbb{1}$ denotes a 2-D tensor with all elements being $1$. $H$ and $N$ are the image height and length of recovered video clip. Instead of directly using the absolute positional encoding of latent frames, we further compute the relative one to $E_R$:
\begin{equation}
	\begin{split}
     \textbf{E} =\{ (E_R - E_{S^t}), t = 0,1,\cdots, N-1 \}
	\end{split}
	\vspace{-1mm}
\end{equation}
Finally, the positional encoding map will be concatenated to $R$ and taken as input to RS branch of MIB.

\begin{table*}[!tp]
	\centering
	\caption{\textbf{Quantitative comparisons} of reconstructed latent frame sequence with lengths of $3$, $5$ and $9$ on RealBR. Subscript of AfB denotes different motion guidance used, while subscripts of RIFE and IFED suggest input settings. `\emph{B}-\emph{R}' is our proposed dual blur-RS view and `\emph{n}$\cdot$\emph{B}' is the setting using \emph{n} neighboring blur frames to tackle motion ambiguity. The performance is measured with mean PSNR, SSIM and LPIPS.
		We also compute the running time, number of parameters and FLOPs.}
	\label{tab:compare_with_sota}
	\resizebox{1\linewidth}{!}
	{
		\begin{tabular}{rccccccccccccc}
			\toprule
			\multirow{2}{*}{Method}& \multirow{2}{*}{Input} & \multicolumn{3}{c}{$\times$3} & \multicolumn{3}{c}{$\times$5} & \multicolumn{3}{c}{$\times$9} & \multirow{2}{*}{\begin{tabular}[c]{@{}l@{}}Time  \\ \quad(s)\end{tabular}} & \multirow{2}{*}{\begin{tabular}[c]{@{}l@{}}Params\\    \quad(M)\end{tabular}} & \multirow{2}{*}{\begin{tabular}[c]{@{}l@{}}FLOPs\\ \quad(G)\end{tabular}} \\ \cmidrule(lr){3-5}\cmidrule(lr){6-8}\cmidrule(lr){9-11}
			
			& & PSNR     & SSIM     & LPIPS    & PSNR    & SSIM    & LPIPS   & PSNR    & SSIM    & LPIPS   &      &        &   \\ \cmidrule(lr){1-11}\cmidrule(lr){12-14}
			
			LEVS~\cite{jin2018learning}& \emph{1}$\cdot$\emph{B} & 21.77 & 0.7042 & 0.2886 & 21.62 &  0.7153 & 0.2683 & 21.83 & 0.7277 & 0.2535 &  1.47 & 15.9 & 304  \\ \cmidrule(lr){1-11}
			
			AfB$_p$ ~\cite{zhong2022animation}& \multirow{4}{*}{\emph{2}$\cdot$\emph{B}}  & 21.50 &  0.7596 & 0.4102 & 21.65 & 0.7648  & 0.4055 & 21.82  & 0.7686 & 0.4017 & \cellcolor{second}0.15  &  190 & 839  \\
			
			AfB$_v$ ~\cite{zhong2022animation}&   & 22.83 &  0.7877 & 0.3904 & 22.96 & 0.7903  & 0.3883 & 23.10  & 0.7924 & 0.3860 & 0.22  & 129 & 793  \\
			
			RIFE$_{B}$~\cite{huang2022real} &   & 24.60  & 0.8172 & 0.2254  & 24.73 & 0.8199 & 0.2268 & 24.83 & 0.8219 & 0.2268 &  1.33   & 54.8 & 71.1    \\
			
			IFED$_{B}$~\cite{zhong2022bringing} &   & 24.45  & 0.8105 & 0.1817  & 24.62 & 0.8141 & 0.1811 & 24.74 & 0.8164 & 0.1798 & 1.33 & \cellcolor{second}10.8  & \cellcolor{best}29.5    \\ \cmidrule(lr){1-11}
			
			BiT ~\cite{zhong2022blur}& \emph{3}$\cdot$\emph{B}  & 21.90 & 0.7664  & 0.2583 & 21.88 & 0.7694 &  0.2574 & 22.02  & 0.7729 & 0.2546 & \cellcolor{best}0.11 &  11.3 & \cellcolor{second}57.4 \\ \cmidrule(lr){1-11}

			DeMFI~\cite{oh2022demfi} & \emph{4}$\cdot$\emph{B} & 25.55  & 0.8485 & 0.2247  & 25.26 & 0.8466 & 0.2275 & 26.20  & 0.8577 & 0.2165 &  4.86   &  \cellcolor{best}7.41  & 420    \\  \cmidrule(lr){1-11}
			
			RIFE$_{BR}$~\cite{huang2022real} & \multirow{3}{*}{\emph{B}-\emph{R}}   & 30.26  & 0.8983 & 0.1071  & 30.53 & 0.9030 & 0.1046 & 30.67 & 0.9053 & 0.1042 &  1.33   & 54.8   & 71.1    \\
			
			IFED$_{BR}$~\cite{zhong2022bringing} &   & \cellcolor{second}30.46  & \cellcolor{second}0.9030 & \cellcolor{second}0.0467  & \cellcolor{second}30.70 & \cellcolor{second}0.9064 & \cellcolor{best}0.0445 & \cellcolor{second}30.84 & \cellcolor{second}0.9084 & \cellcolor{best}0.0434 & 1.33  & \cellcolor{second}10.8   & \cellcolor{best}29.5    \\
			
			Ours &  &  \cellcolor{best}30.87 & \cellcolor{best}0.9073 & \cellcolor{second}0.0696  & \cellcolor{best}31.05  &\cellcolor{best}0.9103  &  \cellcolor{second}0.0684 & \cellcolor{best}31.15 & \cellcolor{best}0.9120 & \cellcolor{second}0.0678 &  1.30  & 105   & 183  \\ \bottomrule
		\end{tabular}
	}
	
\end{table*}

\begin{figure*}[!tb]
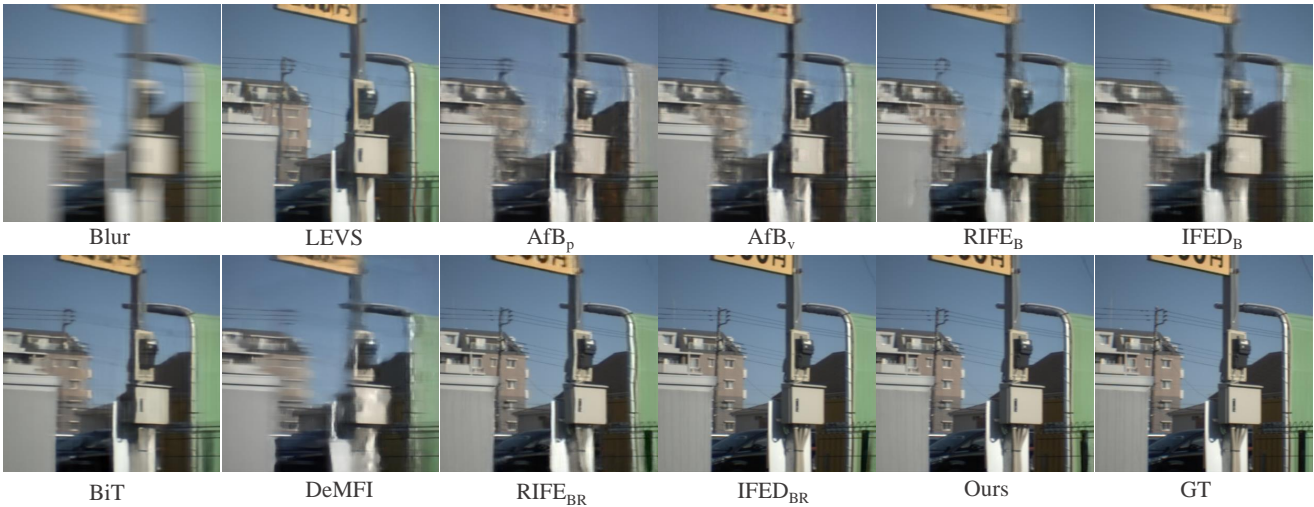

    \vspace{-2mm}
	\centering
	\mfigure{1}{exp/compare_with_sota_v3.pdf}
	\vspace{-4mm}
	\caption{
		\textbf{Qualitative comparison.} Our model outperforms the approaches approximating latent motion fields relying on adjacent blurry inputs. Especially, RIFE$_{BR}$ and IFED$_{BR}$ implemented by dual Blur-RS view reconstruct much sharper details than RIFE$_B$ and IFED$_B$.
	}
	\label{fig:compare_with_sota}
	\vspace{-6mm}
\end{figure*}

\vspace{-2mm}
\paragraph{Shutter Alignment and Aggregation}
SAA module promotes the information propagation across two streams from two perspectives: aggregated and aligned features. The aggregated feature $\phi_{agg}$ is extracted by feeding the concatenation of bidirectional displacement maps to encoder blocks (EB), which address the correlation between two input views, while aligned features $\phi_{alin_B}$ and $\phi_{alin_R}$ focus on absorbing uniquely beneficial parts from each other to stress the contextual characterization and temporal abstraction, respectively. 

The SAA module mainly consists of two encoders implemented by convolutional layers with multi-output strategy like MIMO-UNet~\cite{cho2021rethinking}. Two spatial transformer networks (STN) predict a global transformation conditioned on the output of each encoder to spatially transform features, making the model more robust to visual distortions. We then feed transformed features into their corresponding Conv. blocks to generate bidirectional motion fields. A connection is built through downsampling coarse outputs to next block and the aggregated feature $\phi_{agg}$ is extracted from output of the second encoder. Inference process with $i^{th}$ SAA is as follows:
\begin{equation}
	\begin{split}
		 F^i, M^i= \mathcal{SAA}_i \left(B,R, F^{i-1},M^{i-1} \right)
	\end{split}
	\vspace{-1mm}
\end{equation}


\section{Experiments}
\label{sec:experiments}

As explained in \secref{data_collect}, each blur-RS pair corresponds to 9 high-speed sharp frames. So, for training and validation, each sample comprises a paired input $(B,R)$ and groundtruth video clip $S'$ with a length of 9. Our model is trained by using Adam optimizer~\cite{kingma2014adam} with epoch of $800$. The initial learning rate is set to $10^{-4}$ and decreases to $10^{-6}$ through a cosine annealing scheduler. To augment the training data, we first crop the samples into $512$ and then conduct random horizontal flipping and channel reverse. Experiments are performed on two GPUs of NVIDIA Tesla V100 with batch size of 8.
Besides conducting comparisons on our collected real-world dataset RealBR, we also train all models on synthesized dataset based on GOPRO data~\cite{nah2017deep}. 

\subsection{Comparison with SOTA methods}
\label{comp_sota}
We compare our model with existing state-of-the-arts to handle motion ambiguity of blur decomposition including LEVS~\cite{jin2018learning}, AfB ~\cite{zhong2022animation} and BiT ~\cite{zhong2022blur}. Notably, the AfB is implemented by using different motion guidance: learned by a predictor (AfB$_p$) or extracted from neighboring blur frames (AfB$_v$). In addition, considering that the cutting-edge methods RIFE~\cite{huang2022real} for video frame interpolation and IFED~\cite{zhong2022bringing} for RS temporal super-resolution can be easily adapted to our blur-RS setting, we therefore combine the two models with our setting (denoted as RIFE$_{BR}$ and IFED$_{BR}$) and conduct the experiments. As a contrast, results of these two models using consecutive blur frames (denoted as RIFE$_B$ and IFED$_B$) are also provided. Although, we focus on blur decomposition, we also compared against a blur frame interpolation method, DeMFI~\cite{oh2022demfi} that could be converted to our setting for fair comparison. UTI-VFI~\cite{zhang2020video}, TNTI~\cite{jin2019learning} and BIN~\cite{shen2020blurry}, which cannot distinguish exposure or deadtime when interpolating, are unable to be integrated into our comparison experiments without losing fairness. To better demonstrate the performance of all models, we retrained them on our collected data RealBR.

~\tabref{compare_with_sota} shows quantitative comparisons of all methods on reconstructing video sequence with lengths of $3$, $5$ and $9$. Overall, retrieving a video from single blurred image is quite challenging. The carefully designed supervision of LEVS is barely useful to solve the ambiguity posed by averaging effects. On the other hand, resorting to predicting motion between neighboring inputs, models can indeed speculate the temporal order of latent frames to a certain extent. DeMFI obtains the highest performance among those methods but still lower than our model. Benefited from the dual view, the performance gains of RIFE$_{BR}$ and IFED$_{BR}$, compared with RIFE$_B$ and IFED$_B$, are quite remarkable (at least 5.6 dB on PSNR). The improvement sufficiently demonstrate the effectiveness of blur-RS combination. Reconstructed frames at middle time are presented in~\figref{compare_with_sota} for visual comparison. Although existing methods offer reduced distortions, they do not fully restore local details and structures, whereas our results are significantly clearer. \tabref{compare_with_sota} also presents the computed complexity of all algorithms. Specifically, the FLOPs and running time were evaluated by recovering $9$ latent frames with a size of $256 \times 256$ on an NVIDIA Geforce RTX 3090. The results indicate that our model's complexity is in moderate level.

To better substantiate the ability of our model in motion direction disambiguation and local details recovery, we apply all models to generating $9$ consecutive latent frames, whose visual results are given in supplemental materials. We also provide the video demo for a comprehensive comparisons.

\begin{table}[!h]
	\vspace{-2mm}
	\centering
	\caption{\textbf{Model ablation} on RealBR dataset. `T' denotes temporal encoding and `Single' indicates using single branch. 
	}
	\label{tab:ablation_study}
	\scriptsize
	{
		\begin{tabular}{lccc}
			\toprule
			\textbf{Variants} & \textbf{PSNR} ($\uparrow$)    & \textbf{SSIM} ($\uparrow$)    & \textbf{LPIPS} ($\downarrow$)   \\ \midrule
			W/o T (v1)     &  31.06  &  0.9104  & 0.0690  \\
			W/o SAA (v2)   &  27.96  &  0.8645  & 0.1442 \\
			Single (v3)    &  30.33  &  0.9013  & 0.0929   \\
			Ours (full)    & 31.15 & 0.9120 & 0.0678  \\ \bottomrule
		\end{tabular}
	}
	\vspace{-5mm}
\end{table}

\subsection{Model Ablation}
To assess the efficacy of the components in our proposed model, we conducted an ablation study as depicted in \tabref{ablation_study}. The experiments demonstrate performance of three model variants: $v_1$ (without temporal positional encoding), $v_2$ (without shutter alignment and aggregation), and $v_3$ (using single motion interpretation branch that takes concatenation of blur and RS views as input). Our findings reveal that  dual stream with mutual incentive through SAA module is effective to handle blur decomposition and the temporal positional encoding further improves the performance.

\subsection{Challenging Scenarios}

\begin{figure}[!h]
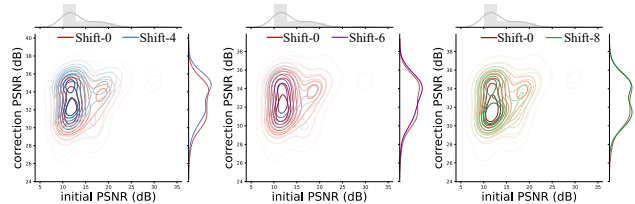

	\vspace{-2mm}
	\mfigure{1}{exp/misaligned_drawing.pdf}
	\vspace{-6mm}
	\caption{
		\textbf{PSNR distribution} of our method with one aligned (`Shift-$0$') and three misaligned views(`Shift-$4$', `Shift-$6$' and `Shift-$8$') under a selected sequence. The horizontal axis is initial PSNR computed by blur view and the first latent frame while the vertical axis denotes PSNR computed between corrected blur view and its ground truth.
	}
	\label{fig:misaligned_views}
\end{figure}

\heading{Misaligned Views}
RS view is taken as a motion guidance and complement of local details to reconstruct multiple latent frames. As discussed in~\cite{zhong2022bringing}, the ambiguity mainly lies in the forward and backward directions of the motions. Hence the motion guidance has no need to be very precise and tolerates some spatial misalignment to blur view. To validate this, we randomly shift RS view in image space along horizontal and vertical axes. The maximal translational offsets are set as $4,6,8$ pixels, respectively. Then we retrained our model on this misaligned pairs and provide comparisons in ~\tabref{compare_with_shift} and~\figref{misaligned_views}. There is no obvious drop, which verifies the robustness of our dual-view setting to misalignment.

\begin{table}[!t]
	\vspace{-4mm}
	\centering
	\caption{Quantitative comparisons on misaligned RS-Blur view. `Shift-$n$' denotes misalignment with maximal offsets $n$ and `Noise-$m$' is an experiment conducted on synthesized low-light RS view under peak value $m$.}
	\label{tab:compare_with_shift}
	\resizebox{\linewidth}{!}
	{
		\begin{tabular}{rccccccccc}
			\toprule
			\multirow{2}{*}{Method} & \multicolumn{3}{c}{$\times$3} & \multicolumn{3}{c}{$\times$5} &  \multicolumn{3}{c}{$\times$9} \\ \cmidrule(lr){2-4}\cmidrule(lr){5-7}\cmidrule(lr){8-10}
			
			& PSNR     & SSIM     & LPIPS    & PSNR    & SSIM    & LPIPS & PSNR    & SSIM    & LPIPS   \\ \midrule
			
			Shift-$4$ & 30.95  & 0.9077 & 0.0705  &  31.18 &  0.9113 & 0.0696 & 31.32 & 0.9135 & 0.0690 \\
			
			Shift-$6$ & 30.82 & 0.9062 & 0.0747 & 31.05 & 0.9098 & 0.0727 & 31.17 & 0.9117 & 0.0724\\		 	
			
			Shift-$8$ & 30.56 & 0.9011 & 0.0907 & 30.73 & 0.9038 & 0.0915 & 30.87 & 0.9065 & 0.0887 \\
   
            Noise-$300$ &30.56 &0.8979 &0.0849 &30.77 &0.9028&0.0841 & 30.88& 0.9053&0.0844 \\
   
			Noise-$500$ &30.92 &0.9012 &0.0849 &30.98 &0.9048&0.0848 & 30.99& 0.9064&0.0848 \\
   
			Noise-$800$ &30.95 &0.9072 &0.0823 &31.04 &0.9083&0.0805 & 31.04& 0.9084&0.0805\\
   
			Ours &   30.87 & 0.9073 & 0.0696  & 31.05  & 0.9103  &  0.0684 & 31.15 & 0.9120 & 0.0678 \\ \bottomrule
		\end{tabular}
	}
	\vspace{-6mm}
\end{table}

\begin{figure}[!ht]
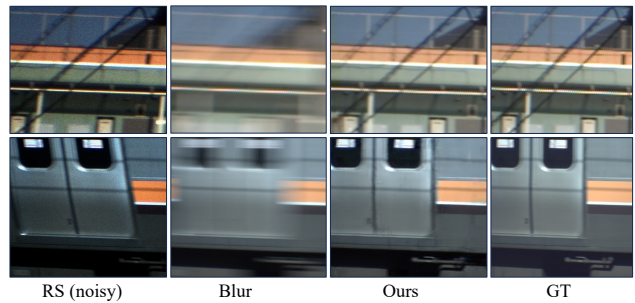

	\vspace{-4mm}
    \mfigure{1}{exp/low_lit_comparison.pdf}
    \vspace{-6mm}
	\caption{
		\textbf{Visual results} of our proposed method under low-lit scenes with peak value $500$. Due to short exposure time of each rows in RS view, it suffers from obvious noise. But our setting is still capable of dealing with this challenge. Best viewed in zoom.
	}
	\label{fig:low_lit_scenes}
	\vspace{-6mm}
\end{figure}

\heading{Low-lit Scenes}
Following the conventional setting, we select proper exposure time of rows for avoiding saturation in GS view and reducing blur in RS view. But it is likely that RS observation will suffer from noise when presenting low-lit scenes. Hence, we further explore effects of noisy RS observations to our method. Following~\cite{lv2018mbllen,lore2017llnet}, we apply a random gamma adjustment and Poisson noise to clear RS view to synthesize low-light captures. Different peak values are chosen to simulate noise of different intensities. The quantitative results are presented in~\tabref{compare_with_shift} showing that low-lit scene causes about $0.27$ dB drop on PSNR, which still outperforms the second-best approach in~\tabref{compare_with_sota}. The visual results are also presented in~\figref{low_lit_scenes}.

\begin{table}[!thp]
	\vspace{-2mm}
	\centering
	\caption{\textbf{Quantitative comparisons} on RS image temporal super-resolution. Subscripts of RIFE and IFED suggests input settings. `\emph{n}$\cdot$\emph{R}' is the setting using \emph{n} neighboring RS frames to tackle initial-state ambiguity.}
	\vspace{-2mm}
	\label{tab:compare_with_rs}
	\resizebox{\linewidth}{!}
	{
		\begin{tabular}{rccccccc}
			\toprule
			\multirow{2}{*}{Method}& \multirow{2}{*}{Input} & \multicolumn{3}{c}{$\times$5} & \multicolumn{3}{c}{$\times$9} \\ \cmidrule(lr){3-5}\cmidrule(lr){6-8}
			
			& & PSNR     & SSIM     & LPIPS    & PSNR    & SSIM    & LPIPS    \\ \midrule
			
			RSSR ~\cite{fan2021inverting}& \multirow{4}{*}{\emph{2}$\cdot$\emph{R}}  & 21.92 &  0.7633 & 0.1526 & 22.82 & 0.7833  & 0.1362   \\ 
			
			CVR ~\cite{fan2022context}&   & 21.70 &  0.7620 & 0.1717 & 22.26 & 0.7761  & 0.1564  \\
			
			RIFE$_{R}$~\cite{huang2022real} &   & 24.14  & 0.8124 & 0.1875  & 24.36 & 0.8181 & 0.1813  \\
			
			IFED$_{R}$~\cite{zhong2022bringing} &   & 24.33  & 0.8033 & 0.1032  & 24.54 & 0.8085 & 0.0989  \\ \midrule
			
			RIFE$_{BR}$~\cite{huang2022real} & \multirow{3}{*}{\emph{B}-\emph{R}}   & 30.53 & 0.9030 & 0.1046 & 30.67 & 0.9053 & 0.1042    \\
			
			IFED$_{BR}$~\cite{zhong2022bringing} &   & \cellcolor{second}30.70 & \cellcolor{second}0.9064 & \cellcolor{best}0.0445 & \cellcolor{second}30.84 & \cellcolor{second}0.9084 & \cellcolor{best}0.0434  \\
			
			Ours &  &  \cellcolor{best}31.05  & \cellcolor{best}0.9103  & \cellcolor{second}0.0684 & \cellcolor{best}31.15 & \cellcolor{best}0.9120 & \cellcolor{second}0.0678  \\ \bottomrule
		\end{tabular}
	}
	\vspace{-4mm}
\end{table}

\begin{figure}[!t]
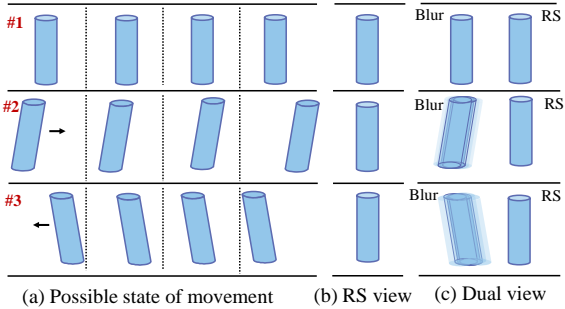

	\centering
	\resizebox{0.95\linewidth}{!}
	{
		\mfigure{1}{exp/initial_state_ambiguity_v2.pdf}
	}
	\vspace{-2mm}
	\caption{
		\textbf{Initial-state ambiguity of RS observation.} In this toy example, we show one object moving horizontally. (a) shows three possible motion states (static, moving right and moving left) during the exposure time. (b) presents corresponding RS observations. They are all identical due to different initial-state (upright, tilted to the right, and tilted to the left), which brings about ambiguity to RS temporal super-resolution. (c) In our dual Blur-RS setting, blur view sufficiently indicated initial-state of latent frames.
	}
	\label{fig:initail_state_ambiguity}
	\vspace{-2mm}
\end{figure}

\subsection{Justification for Dual Blur-RS Setting}
\label{sec:setting_justification}
Our proposed dual blur-RS setting requires extra view compared with existing methods using neighboring frames, which may increase the cost of device. But considering the performance gains, it is worthwhile. The reason is that, on one hand, blur decomposition entails motion ambiguity as shown in ~\figref{motion_ambiguity}. The quantitative and qualitative comparisons with SOTA methods corroborate that RS view can guide blur input to infer a temporally plausible video sequence with more local details.

On the other hand, we delve deeper into this blur-RS setting under the context of RS temporal super resolution to well justify its superiority. Although human's visual perception to RS effects is not that sensitive like blur, the high ill-posedness makes the correction process barely tractable even harder than debluring~\cite{Ji_2023_ICCV}. One possible reason is the initial-sate ambiguity as illustrated in ~\figref{initail_state_ambiguity}. Given an RS observation of upright cylinder, due to the unknown of its initial state, there exists three possible motion patterns within exposure. Similarly, ~\cite{fan2021inverting,fan2022context} exploit consecutive RS inputs to mitigate the ambiguity, while ~\cite{zhong2022bringing} resorts to dual RS view with reversed exposure direction. In ~\tabref{compare_with_rs}, we compare our solution with corresponding SOTA methods from RS temporal super resolution. The RIFE and IFED are also implemented by taking neighboring RS frames as input, denoted as RIFE$_R$ and IFED$_R$ respectively. All experiments validated that blur view provide cues of initial state and global context to RS counterpart that effectively boost reconstruction performance.

\begin{figure*}[!ht]
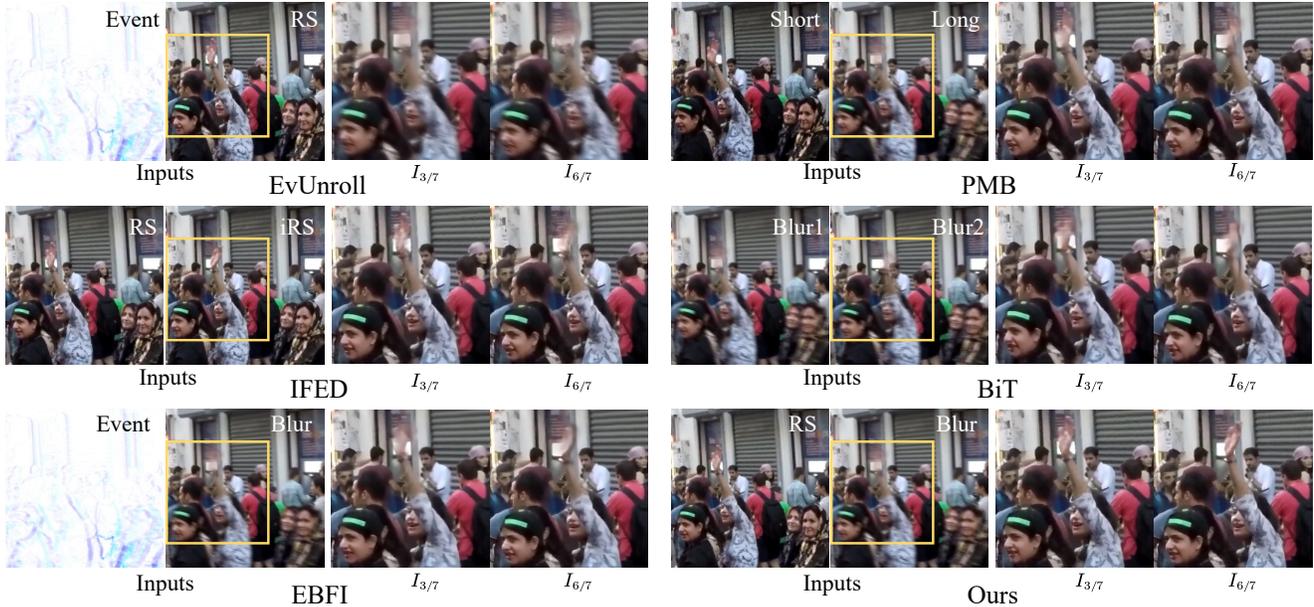

	\mfigure{1}{exp/compare_settings2.pdf}
	\vspace{-4mm}
	\caption{
		\textbf{Visual results} of comparisons with competitive settings on synthetic data. Best viewed in zoom.
	}
	\label{fig:compare_on_synthesic}
	\vspace{-2mm}
\end{figure*}

 Notably, because of lacking different modal data in RealBR, comparisons between more competitive settings: IFED~\cite{zhong2022bringing} with \emph{dual reversed RS views}, EvUnroll~\cite{zhou2022evunroll} and EBFI~\cite{weng2023event} assisted by \emph{event camera}, PMB~\cite{rengarajan2020photosequencing} using \emph{short and long exposure}, are conducted on synthetic data. The details of data synthesis are included in supplementary. From~\tabref{compare_with_rs_synthesic} and~\figref{compare_on_synthesic}, we further validate that our setting and method have superiority against existing solutions.

\begin{table}[!tp]
	\centering
	\caption{\textbf{Quantitative comparisons} with more competitive settings under task of blur decomposition and RS temporal super resolution based on synthetic data. Dual reserved RS setting is denoted as `\emph{R}-\emph{iR}'. `\emph{B}-\emph{Event}' and `\emph{R}-\emph{Event}' are blur decomposition and RS temporal super resolution assisted by event camera, respectively. `\emph{B}-\emph{SL}' means photosequencing from blur using short long exposure.}
	\label{tab:compare_with_rs_synthesic}
	\resizebox{\linewidth}{!}
	{
		\begin{tabular}{rccccccc}
			\toprule
			\multirow{2}{*}{Method}& \multirow{2}{*}{Input} & \multicolumn{3}{c}{$\times$3} & \multicolumn{3}{c}{$\times$7} \\ \cmidrule(lr){3-5}\cmidrule(lr){6-8}
			
			& & PSNR     & SSIM     & LPIPS    & PSNR    & SSIM    & LPIPS    \\ \midrule
			
			LEVS~\cite{jin2018learning} & \emph{1}$\cdot$\emph{B} & 17.27 & 0.6063 & 0.3410 & 16.64 & 0.58 & 0.3811  \\ \midrule
			
			AfB$_p$ ~\cite{zhong2022animation}& \multirow{4}{*}{\emph{2}$\cdot$\emph{B}}  & 23.38 &  0.7411 & 0.2271 & 23.41  & 0.7517 & 0.2183   \\
			AfB$_v$ ~\cite{zhong2022animation}&   & 28.10 &  0.8760 & 0.1496 & 28.39 & 0.8815 & 0.1461 \\
			
			RIFE$_{B}$~\cite{huang2022real} &    & 31.26 & 0.9410 & 0.0896  & 31.49 & 0.9430 & 0.0892    \\
			
			IFED$_{B}$~\cite{zhong2022bringing} &   & 29.46 &  0.9193 & 0.0897 & 29.75 & 0.9225 & 0.0874 \\ \midrule
			
			BiT ~\cite{zhong2022blur}& \emph{3}$\cdot$\emph{B}  & 32.31  & 0.9234 & 0.0708 & 32.56  & 0.9266 & 0.0691 \\ \midrule
			
			DeMFI~\cite{oh2022demfi} & \emph{4}$\cdot$\emph{B} & 27.57  & 0.9002 & 0.1332 & 27.44  & 0.8984 & 0.1304   \\  \midrule
			
			PMB~\cite{rengarajan2020photosequencing} & \emph{B}-\emph{SL} & \cellcolor{best}35.48 & \cellcolor{second}0.9723  & 0.0349 & \cellcolor{second}35.11 & 0.9715  & 0.0324  \\  \midrule
			
			EBFI~\cite{weng2023event}& \emph{B}-\emph{Event} & 33.21 & 0.9568  & 0.0703 & 33.51 & 0.9591   & 0.0685  \\ 
			
			\bottomrule
			
			RSSR ~\cite{fan2021inverting}& \multirow{4}{*}{\emph{2}$\cdot$\emph{R}}  & 22.73 &  0.8116 & 0.1039 & 22.65 & 0.8090  & 0.1154   \\ 
			
			CVR ~\cite{fan2022context}&   & 23.50 &  0.8342 & 0.0818 & 23.47 & 0.8332  & 0.0815  \\ 
			
			RIFE$_{R}$~\cite{huang2022real} &   & 24.16 & 0.8318 & 0.1697   & 24.32 & 0.8365 & 0.1618 \\
			
			IFED$_{R}$~\cite{zhong2022bringing}& & 28.30 & 0.9122  & 0.0475 & 28.63 & 0.9181  &0.0446   \\      \midrule
			
			IFED~\cite{zhong2022bringing}& \emph{R}-\emph{iR} & 30.89 & 0.9417  & 0.0372 & 31.96 & 0.9530  &0.0307 \\ \midrule
			
			EvUnroll~\cite{zhou2022evunroll}& \emph{R}-\emph{Event} & 33.06 &  0.9558 & 0.0737 & 33.48 & 0.9587  &0.0699   \\ 
			
			\bottomrule
			
			RIFE$_{BR}$~\cite{huang2022real} &  \multirow{3}{*}{\emph{B}-\emph{R}}  & 34.49 & 0.9701 & 0.0398   & 35.02 & \cellcolor{second}0.9733 & 0.0366    \\
			
			IFED$_{BR}$~\cite{zhong2022bringing} &   & 33.03  & 0.9627 & \cellcolor{second}0.0332 & 33.72 & 0.9675 &  \cellcolor{best}0.0304    \\
			
			Ours &  & \cellcolor{second}34.92 & \cellcolor{best}0.9732  & \cellcolor{best}0.0310  & \cellcolor{best}35.51 & \cellcolor{best}0.9764 & \cellcolor{second}0.0305    \\ \bottomrule
		\end{tabular}
	}
	\vspace{-3mm}
\end{table}

\section{Conclusion}
\label{sec:conclusion}

In this paper, we have proposed a novel cross-shutter setting for motion decomposition of a single blurry image, inspired by the complementary exposure characteristics of GS and RS cameras. Since this setting is new, we first developed a triaxial image capture system to collect triplets of blurry image, rolling shutter image and consecutive sharp frames at higher frame rate. In the arithmetic aspect, we proposed a novel network architecture that
actively addresses  the contextual characterization and temporal abstraction in a mutual incentive manner. Experiments on our real dataset have verified the effectiveness of proposed algorithm. With synthetic data, we further demonstrated the superiority of our global-shutter/rolling-shutter dual imaging setting. Our current implementation requires a beamsplitter to align two different shutters, which is demanding for compact mobile devices, and our future work is to explore the feasibility of using synchronized sensors placed in parallel.

{
    \small
    \bibliographystyle{ieeenat_fullname}
    \bibliography{main}
}

\end{document}